%% file: Chapter.tex
\documentclass{llncs}

\usepackage{float}
\usepackage{placeins}
\usepackage{graphicx}
\usepackage{color}
\usepackage{tikz}
\usepackage{caption}
\usetikzlibrary{arrows}
\usepackage{wrapfig}
\usepackage{subfig}
\usepackage{pbox}
\usepackage{rotating}
\usepackage{pdflscape}
\usepackage{w-bookps}
\usepackage{adjustbox}
\usepackage{natbib}
\usepackage{chapterbib}
\usepackage{amsmath}
\usepackage{caption}

\makeatletter
\renewcommand\tagform@[1]{\maketag@@@ {[\ignorespaces #1\unskip \@@italiccorr ]}}
\makeatother

\begin{document}

\title{Twitter Sentiment Analysis: Lexicon Method, Machine Learning Method and Their Combination}

\author{Olga Kolchyna\inst{1}, Th\'{a}rsis T. P. Souza\inst{1}, Philip C. Treleaven\inst{1}\inst{2} \and Tomaso Aste\inst{1}\inst{2}}

\authorrunning{Kolchyna et al.} % abbreviated author list (for running head)
%
%%%% list of authors for the TOC (use if author list has to be modified)
\tocauthor{Kolchyna et al.}
\institute{Department of Computer Science, UCL, Gower Street, London, UK,\and
Systemic Risk Centre, London School of Economics and Political Sciences, London, UK}

\maketitle

\begin{abstract} 
This paper covers the two approaches for sentiment analysis: i) lexicon based method; ii) machine learning method. We describe several techniques to implement these approaches and discuss how they can be adopted for sentiment classification of Twitter messages. We present a comparative study of different lexicon combinations and show that enhancing sentiment lexicons with emoticons, abbreviations and social-media slang expressions increases the accuracy of lexicon-based classification for Twitter. We discuss the importance of feature generation and feature selection processes for machine learning sentiment classification. To quantify the  performance of the main sentiment analysis methods over Twitter we run these algorithms on a benchmark Twitter dataset from the SemEval-2013 competition, task 2-B. The results show that machine learning method based on SVM and Naive Bayes classifiers outperforms the lexicon method. We present a new ensemble method that uses a lexicon based sentiment score as input feature for the machine learning approach. The combined method proved to produce more precise classifications. We also show that employing a cost-sensitive classifier for highly unbalanced datasets yields an improvement of sentiment classification performance up to 7\%. 
\end{abstract}

\smallskip


\newpage

\section{Introduction}

Sentiment analysis is an area of research that investigates people's opinions towards different matters: products, events, organisations \citep{Bing:2012}. The role of sentiment analysis has been growing significantly with the rapid spread of social networks, microblogging applications and forums. Today, almost every web page has a section for the users to leave their comments about products or services, and share them with friends on Facebook, Twitter or Pinterest - something that was not possible just a few years ago. Mining this volume of opinions provides information for understanding collective human behaviour and it is of valuable commercial interest. For instance, an increasing amount of evidence points out that by analysing sentiment of social-media content it might be possible to predict the size of the markets \citep{TwitterPredicts} or unemployment rates over time \citep{Unemployment-2014}. 

One of the most popular microblogging platforms is Twitter. It has been growing steadily for the last several years and has become a meeting point for a diverse range of people: students, professionals, celebrities, companies and politicians. This popularity of Twitter results in the enormous amount of information being passed through the service, covering a wide range of topics from people well-being to the opinions about the brands, products, politicians and social events. In this contexts Twitter becomes a powerful tool for predictions. For example, \citep{Asur:2010} was able to predict from Twitter analytics the amount of ticket sales at the opening weekend for movies with 97.3\% accuracy, higher than the one achieved by the  Hollywood Stock Exchange, a known prediction tool for the movies. 

In this paper, we present a step-by-step approach for two main methods of sentiment analysis: lexicon based approach \citep{Taboada:2011:LMS:2000517.2000518}, \citep{Ding:2008:HLA:1341531.1341561} and machine learning approach \citep{PAK10.385}. We show that accuracy of the sentiment analysis for Twitter can be improved by combining the two approaches: during the first stage a lexicon score is calculated based on the polarity of the words which compose the text, during the second stage a machine learning model is learnt that uses the lexicon score as one of the features. The results showed that the combined approach outperforms the two approaches. We demonstrate the use of our algorithm on a dataset from a  popular Twitter sentiment competition SemEval-2013, task 2-B \citep{SemEval-2013}. In \citep{Souza:2015} our algorithm for sentiment analysis is also successfully applied to 42,803,225 Twitter messages related to companies from the retail sector to predict the stock price movements.

\input sec-sentiment

\section{Conclusion}

In this  paper  we have presented  the  review  of  two main approaches for sentiment analysis, a lexicon based method and a machine learning method. 

In the lexicon based approach we compared the performance of three lexicons: i) an Opinion lexicon (OL); ii) an Opinion lexicon enhanced with manually created corpus of emoticons, abbreviations and social-media slang expressions (OL + EMO); iii) OL + EMO further enhanced with automatically generated lexicon (OL + EMO + AUTO). We showed that on a benchmark Twitter dataset, OL + EMO lexicon outperforms both, the traditional OL and a larger OL + EMO + AUTO lexicon. These results demonstrate the importance of incorporating expressive signals such as emoticons, abbreviations and social-media slang phrases into lexicons for Twitter analysis. The results also show that larger lexicons may yield a decrease in performance due to ambiguity of words polarity and increased model complexity (agreeing with \citep{Ghiassi:2013:TBS:2506578.2506879}).

In the machine learning approach we propose to use a lexicon sentiment obtained during the lexicon based classification as an input feature for training classifiers. The ranking of all features based on the information gain scores during the feature selection process revealed that the lexicon feature appeared on the top of the list, confirming its relevance in sentiment classification. We also demonstrated that in case of highly unbalanced datasets the utilisation of cost-sensitive classifiers improves accuracy of class prediction: on the benchmark Twitter dataset a cost-sensitive SVM yielded 7\% increase in performance over a standard SVM.

\section*{Acknowledgments}

We thank the valuable feedback from the two anonymous reviewers. T.A. acknowledges support of the UK Economic and Social Research Council (ESRC) in funding the Systemic Risk Centre (ES/K002309/1). O.K. acknowledges support from the company Certona Corporation. T.T.P.S. acknowledges financial support from CNPq - The Brazilian National Council for Scientific and Technological Development. 

\bibliographystyle{apalike}
\bibliography{chapter}

\end{document}

%% file: sec-sentiment.tex
\section{Sentiment Analysis Methodology: Background}
\label{sec:sent}

The field of text categorization was initiated long time ago \citep{Salton:1983}, however categorization based on sentiment was introduced more recently in \citep{opinion.stock.2001, Morinaga:2002:MPR:775047.775098, Pang:2002:TUS:1118693.1118704, Tong2001, Turney:2002:TUT:1073083.1073153, Wiebe:2000:LSA:647288.721121}.

The standard approach for text representation \citep{Salton:1983} has been the bag-of-words mehod (BOW). According to the BOW model, the document is represented as a vector of words in Euclidian space where each word is independent from others. This bag of individual words is commonly called a collection of unigrams. The BOW is easy to understand and allows to achieve high performance (for example, the best results of multi-lable categorization for the Reuters-21578 dataset were produced using BOW approach \citep{288651, Weiss:1999:MTP:630307.630474}). 

The main two methods of sentiment analysis, lexicon-based method (unsupervised approach) and machine learning based method (supervised approach), both rely on the bag-of-words. In the machine learning supervised method the classifiers are using the unigrams or their combinations (N-grams) as features. In the lexicon-based method the unigrams which are found in the lexicon are assigned a polarity score, the overall polarity score of the text is then computed as sum of the polarities of the unigrams.

When deciding which lexicon elements of a message should be considered for sentiment analysis, different parts-of-speech were analysed \citep{PAK10.385, conf/icwsm/KouloumpisWM11}. Benamara et al. proposed the Adverb-Adjective Combinations (AACs) approach that demonstrates the use of adverbs and adjectives to detect sentiment polarity \citep{Benamara07}. In recent years the role of emoticons has been investigated  \citep{conf/aiia/PozziFMB13, Hogenboom:2013:EES:2480362.2480498, Liu20121678, DBLP:conf/kdd/ZhaoDWX12}. In their recent study \citep{Fersini2015} further explored the use of (i) adjectives, (ii) emoticons, emphatic and onomatopoeic expressions and (iii) expressive lengthening as expressive signals in sentiment analysis of microblogs. They showed that the above signals can enrich the feature space and improve the quality of sentiment classification.

Advanced algorithms for sentiment analysis have been developed (see \citep{conf/anlp/Jacobs92, vapnik98statlearn, basili00languagesensitive, schapire00boostexter}) to take into consideration not only the message itself, but also the context in which the message is published, who is the author of the message, who are the friends of the author, what is the underlying structure of the network. For instance, \citep{Hu:2013:ESR:2433396.2433465} investigated how social relations can help sentiment analysis by introducing a Sociological Approach to handling Noisy and short Texts (SANT), \citep{Zhu:2014:TGC:2588555.2593682} showed that the quality of sentiment clustering for Twitter can be improved by joint clustering of tweets, users, and features. In the work by \citep{conf/aiia/PozziMFM13} the authors looked at friendship connections and estimated user polarities about a given topic by integrating post contents with approval relations. Quanzeng You and Jiebo Luo improved sentiment classification accuracy by adding a visual content in addition to the textual information \citep{You:2013:TSI:2501217.2501220}. Aisopos et al. significantly increased the accuracy of sentiment classification by using content-based features along with context-based features \citep{Aisopos:2012:CVC:2309996.2310028}. Saiff et al. achieved improvements by growing the feature space with semantics features \citep{Saif:2012:SSA:2426516.2426549}.

While many research works focused on finding the best features, some efforts have been made to explore new methods for sentiment classification. Wang et al. evaluated the performance of ensemble methods (Bagging, Boosting, Random Subspace) and empirically proved that ensemble models can produce better results than the base learners \citep{Wang201477}. Fersini et al. proposed to use Bayesian Model Averaging ensemble method which outperformed both traditional classification and ensemble methods \citep{Fersini201426}. Carvalho et al. employed genetic algorithms to find subsets of words from a set of paradigm words that led to improvement of classification accuracy \citep{Carvalho:2014:SEA:2682648.2682836}. 

\section{Data Pre-processing for Sentiment Analysis} \label{sec:Preprocessing}
Before applying any of the sentiment extraction methods, it is a common practice to perform data pre-processing. Data pre-processing allows to produce higher quality of text classification and reduce the computational complexity. Typical pre-processing procedure includes the following steps:

\textbf{Part-of-Speech Tagging (POS).} The process of part-of-speech tagging allows to automatically tag each word of text in terms of which part of speech it belongs to: noun, pronoun, adverb, adjective, verb, interjection, intensifier, etc. The goal is to extract patterns in text based on analysis of frequency distributions of these part-of-speech. The importance of part-of-speech tagging for correct sentiment analysis was demonstrated by \citep{manning1999foundations}. Statistical properties of texts, such as  adherence to Zipf’s law can also be used \citep{Piantadosi:2014}. Pak and Paroubek analysed the distribution of POS tagging  specifically for Twitter messages and identified multiple patterns \citep{PAK10.385}. For instance, they found that subjective texts (carrying the sentiment) often contain more pronouns, rather than common and proper nouns; subjective messages often use past simple tense and contain many verbs in a base form and many modal verbs.

There is no common opinion about whether POS tagging improves the results of sentiment classification. Barbosa and Feng reported positive results using POS tagging \citep{Barbosa:2010:RSD:1944566.1944571}, while \citep{conf/icwsm/KouloumpisWM11} reported a decrease in performance.
	
\textbf{Stemming and lemmatisation}. Stemming is a procedure of replacing words with their stems, or roots. The dimensionality of the BOW is reduced when root-related words, such as ``read'', ``reader'' and ``reading'' are mapped into one word ``read''. However, one should be careful when applying stemming, since it might increase bias. For example, the biased effect of stemming appears when merging distinct words ``experiment'' and ``experience'' into one word ``exper'', or when words which ought to be merged together (such as ``adhere'' and ``adhesion'') remain distinct after stemming. These are examples of over-stemming and under-stemming errors respectively. Overstemming lowers precision and under-stemming lowers recall. The overall impact of stemming depends on the dataset and stemming algorithm. The most popular stemming algorithm is Porter stemmer \citep{Porter1980}.

\textbf{Stop-words removal}. Stop words are words which carry a connecting function in the sentence, such as prepositions, articles, etc. \citep{Salton:1983}. There is no definite list of stop words, but some search machines, are using some of the most common, short function words, such as ``the'', ``is'', ``at'', ``which'' and ``on''. These words can be removed from the text before classification since they have a high frequency of occurrence in the text, but do not affect the final sentiment of the sentence.
 
\textbf{Negations Handling. } Negation refers to the process of conversion of the sentiment of the text from positive to negative or from negative to positive by using special words: \textit{``no'',``not'',``don't''} etc. These words are called negations. 
The example of some negation words is presented in the Table \ref{tb:NegationWords}.

\begin{table}[ht]
\caption{Example Negation Words} %title of the table
\label{tb:NegationWords}
\centering
% centering table
\begin{tabular}{c c c c}
\hline
hardly  & cannot & shouldn't & doesnt\\
lack & daren't & wasn't & didnt\\
lacking & don't & wouldn't & hadnt\\
lacks & doesn't & weren't & hasn't\\
neither & didn't & won't & havn't\\
nor & hadn't & without & haven't\\
\hline
\end{tabular}
\label{tab:nonlin}
\end{table}

Handling negation in the sentiment analysis task is a very important step as the whole sentiment of the text may be changed by the use of negation. It is important to identify the scope of negation (for more information see \citep{Councill:2010:WGW:1858959.1858969}). The simplest approach to handle negation is to revert the polarity of all words that are found between the negation and the first punctuation mark following it. For instance, in the text ``I don't want to go to the cinema'' the polarity of the whole phrase ``want to got to the cinema'' will be reverted. 

Other researches introduce the concept of contextual valence shifter \citep{polanyi2006}, which consists of negation, intensifier and diminisher.  Contextual valence shifters have an impact of flipping the polarity, increasing or decreasing the degree to which a sentimental term is positive or negative.

\textbf{But-clauses}. The phrases like \textit{``but'', ``with the exception of'', ``except that'', ``except for''} generally change the polarity of the part of the sentence following them. In order to handle these clauses the opinion orientation of the text before and after these phrases should be set opposite to each other. For example, without handling the \textit{``but-type clauses''} the polarity of the sentence may be set as following:\textit{``I don’ like[-1] this mobile, but the screen has high[0] resolution''}. When ``but-clauses'' is processed, the sentence polarity will be changed to: ``I don't like[-1] this mobile, but the screen has high[+1] resolution''. Notice, that even neutral adjectives will obtain the polarity that is opposite to the polarity of the phrase before the ``but-clause''.

However, the solution described above does not work for every situation. For example, in the sentence ``Not only he is smart, but also very kind'' - the word ``but'' does not carry contrary meaning and reversing the sentiment score of the second half of the sentence would be incorrect. These situations need to be considered separately.
	
\textbf{Tokenisation into N-grams.} Tokenisation is a process of creating a bag-of-words from the text. The incoming string gets broken into comprising words and other elements, for example URL links. The common separator for identifying individual words is whitespace, however other symbols can also be used. Tokenisation of social-media data is considerably more difficult than tokenisation of the general text since it contains numerous emoticons, URL links, abbreviations that cannot be easily separated as whole entities. 

It is a general practice to combine accompanying words into phrases or n-grams, which can be unigrams, bigrams, trigrams, etc. Unigrams are single words, while bigrams are collections of two neighbouring words in a text, and trigrams are collections of three neighbouring words. N-grams  method can decrease bias, but may increase statistical sparseness. It has been shown that the use of n-grams can improve the quality of text classification \citep{Raskutti:2001:SOF:645328.650032, Zhang03questionclassification, Diederich:2003:AAS:776973.776982}, however there is no unique solution for the the size of n-gram. Caropreso et al. conducted an experiment of text categorization on the Reuters-21578 benchmark dataset \citep{Caropreso:2001:LEU:374247.374254}. They reported that in general the use of bigrams helped to produce better results than the use of unigrams, however while using Rocchio classifier \citep{rocchio71relevance} the use of bigrams led to the decrease of classification quality in 28 out of 48 experiments. Tan et al. reported that use of bigrams on Yahoo-Science dataset \citep{Tan02theuse}  allowed to improve the performance of text classification using Naive Bayes  classifier from 65\% to 70\% break-even point, however, on Reuters-21578 dataset the increase of accuracy was not significant. Conversely, trigrams were reported to generate poor performances \citep{PAK10.385}.

\section{Sentiment Computation with Lexicon-Based Approach} \label{sec:Lexicon}

\begin{wrapfigure}{R}{0.43\textwidth}
\centering
\includegraphics[width=0.42\textwidth]{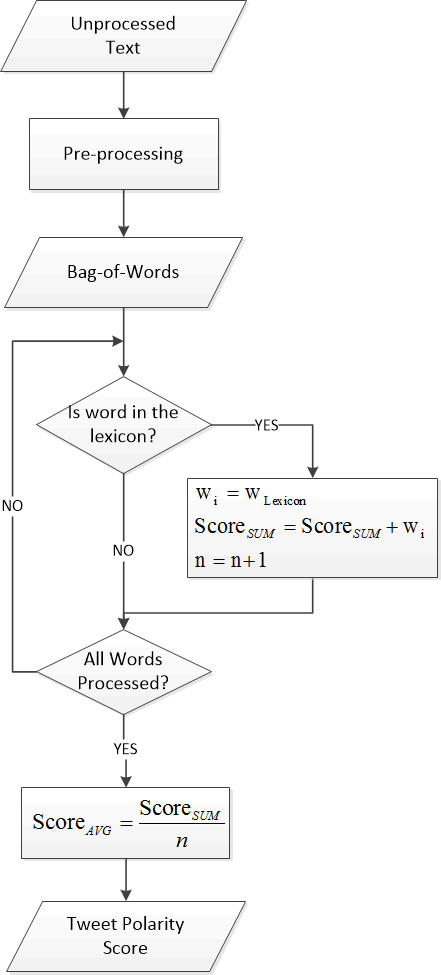}
\caption{\label{fig:SentScore}Schematic representation of methodology for the sentiment calculation.}
\end{wrapfigure}

Lexicon-based approach calculates the sentiment of a given text from the polarity of the words or phrases in that text \citep{Turney:2002:TUT:1073083.1073153}. For this method a lexicon (a dictionary) of words with assigned to them polarity is required. Examples of the existing lexicons include: Opinion Lexicon \citep{BingLexicon}, SentiWordNet \citep{Esuli06}, AFINN Lexicon \citep{AFINN}, LoughranMcDonald Lexicon, NRC-Hashtag \citep{NRC}, General Inquirer Lexicon\footnote{http://www.wjh.harvard.edu/~inquirer/} \citep{Stone:1963}.

The sentiment score $Score$ of the text $T$ can be computed as the average of the polarities conveyed by each of the words in the text. The methodology for the sentiment calculation is schematically illustrated in Figure \ref{fig:SentScore} and can be described with the following steps:

\begin{itemize}

\item \textbf{Pre-processing.} The text undergoes pre-processing steps that were described in the previous section: POS tagging, stemming, stop-words removal, negation handling, tokenisations into N-grams. The outcome of the pre-processing is a set of tokens or a bag-of-words.

\item \textbf{Checking each token for its polarity in the lexicon.} Each word from the bag-of-words is compared against the lexicon. If the word is found in the lexicon, the polarity $w_i$ of that word is added to the sentiment score of the text. If the word is not found in the lexicon its polarity is considered to be equal to zero.

\item \textbf{Calculating the sentiment score of the text.} After assigning polarity scores to all words comprising the text, the final sentiment score of the text is calculated  by dividing the sum of the scores of  words caring the sentiment by the number of such words:

\begin{equation}
Score_{AVG}= \frac{1}{m}\sum_{i=1}^mW_i.
\label{eq:LexiconScore} 
\end{equation}

\item[] The averaging of the score allows to obtain a value of the sentiment score in the range between -1 and 1, where 1 means a strong positive sentiment, -1 means a strong negative sentiment and 0 means that the text is neutral. For example, for the text:

``\textit{A masterful[+0.92] film[0.0] from a master[+1] filmmaker[0.0], unique[+1] in its deceptive[0.0] grimness[0.0], compelling[+1] in its fatalist[-0.84] world[0.0] view[0.0].''}

the sentiment score is calculated as follows:

\[
Score_{AVG}= \frac{0.92 + 0.0 + 1 + 0.0 + 1 + 0.0 + 0.0 + 1 - 0.84 + 0.0 + 0.0}{5} = 0.616.
\]

The sentiment score of 0.616 means that the sentence expresses a positive opinion.

\end{itemize}

The quality of classification highly depends on the quality of the lexicon. Lexicons can be created using different techniques:

\textbf{Manually constructed lexicons}. The straightforward approach, but also the most time consuming, is to manually construct a lexicon and tag words in it as positive or negative. For example, \citep{opinion.stock.2001} constructed their lexicon by reading several thousands of messages and manually selecting words, that were carrying sentiment. They then used a discriminant function to identify words from a training dataset, which can be used for sentiment classifier purposes. The remained words were ``expanded'' to include all potential forms of each word into the final lexicon. Another example of hand-tagged lexicon is The Multi-Perspective-Question-Answering (MPQA) Opinion Corpus\footnote{available at nrrc.mitre.org/NRRC/publications.htm} constructed by \citep{wiebeetal2005a}. MPQA is publicly available and consists of 8,222 subjective expressions along with their POS-tags, polarity classes and intensity. 

Another resource is The SentiWordNet created by \citep{Esuli06}. SentiWordNet  extracted words from WordNet\footnote{http://wordnet.princeton.edu/} and gave them  probability of belonging to positive, negative or neutral classes, and subjectivity score. Ohana and Tierney  demonstrated that SentiWordNet can be used as an important resource for sentiment calculation \citep{ohana2009sentiment}.

\textbf{Constructing a lexicon from trained data}. This approach belongs to the category of the supervised methods, because a training dataset of labelled sentences is needed. With this method the sentences from the training dataset get tokenised and a bag-of-words is created. The words are then filtered to exclude some parts-of-speech that do not carry sentiment, such as prepositions, for example. The prior polarity of words is calculated according to the occurrence of each word in positive and negative sentences. For example, if a word ``success'' is appearing more often in the sentences labelled as positive in the training dataset, the prior polarity of this word will be assigned a positive value.  

\textbf{Extending a small lexicon using bootstrapping techniques}.
Hazivassiloglou and McKeown proposed to extend a small lexicon comprised of adjectives by adding new adjectives which were conjoined with the words from the original lexicon \citep{hatzivassilogloumckeown1997}. The technique is based on the syntactic relationship between two adjectives conjoined with the ``AND'' – it is established that ``AND'' usually joins words with the same semantic orientation. Example: 

``The weather yesterday was nice and inspiring''

Since words ``nice'' and ``inspiring'' are conjoined with ``AND'', it is considered that both of them carry a positive sentiment. If only the word ``nice'' was present in the lexicon, a new word ``inspiring'' would be added to the lexicon. Similarly, \citep{hatzivassilogloumckeown1997} and \citep{kimhovy2004} suggested to expand a small manually constructed lexicon with synonyms and antonyms obtained from NLP resources such as WordNet\footnote{https://wordnet.princeton.edu/}. The process can be repeated iteratively until it is not possible to find new synonyms and antonyms. Moilanen and Pulman also created their lexicon by semi-automatically expanding WordNet2.1 lexicon \citep{Moilanen:Sentiment}. Other approaches include extracting polar sentences by using structural clues from HTML documents \citep{conf/emnlp/KajiK07}, recognising opinionated text based on the density of other clues in the text \citep{Wiebe02}. After the application of a bootstrapping technique it is important to conduct a manual inspection of newly added words to avoid errors.

\section{A Machine Learning Based Approach} \label{sec:ML}

A Machine Learning Approach for text classification is a supervised algorithm that analyses data that were previously labelled as positive, negative or neutral; extracts features that model the differences between different classes, and infers a function, that can be used for classifying new examples unseen before. In the simplified form, the text classification task can be described as follows: given a dataset of labelled data $T_{train}=\{\left(t_1,\ l_1\right),\ \dots ,\ (t_n,\ l_n)\}$ , where each text $t_i$ belongs to a dataset $T$ and the label $l_i$ is a pre-set class within the group of classes \textit{L}, the goal is to  build a learning algorithm that will receive as an input the training set $T_{train}$ and will generate a model that will accurately classify unlabelled texts. 

The most popular learning algorithms for text classification are Support Vector Machines (SVMs) \citep{Cortes:1995:SN:218919.218929, Vapnik:1995:NSL:211359}, Naive Bayes \citep{NaiveBayesSent}; Decision Trees \citep{Mitchell:1996}. Barbosa et al. reports better results for SVMs \citep{Barbosa:2010:RSD:1944566.1944571} while Pak et al. obtained better results for Naive Bayes \citep{PAK10.385}. In the  work  by  \citep{288651} a  decision tree classifier was shown to perform nearly as well as an SVM classifier.

In terms of the individual classes, some researches \citep{Pang:2002:TUS:1118693.1118704} classified texts only as positive or negative, assuming that all the texts carry an opinion. Later \citep{wilsonetal2005b}, \citep{PAK10.385} and \citep{Barbosa:2010:RSD:1944566.1944571} showed that short messages like tweets and blogs comments often just state facts. Therefore, incorporation of the  neutral class into the classification process is necessary.

The process of machine learning text classification can be broken into the following steps:

\begin{enumerate}

\item \textbf{Data Pre-processing.} Before training the classifiers each text needs to be pre-processed and presented as an array of tokens. This step is performed according to the process described in section \ref{sec:Preprocessing}. 

\item \textbf{Feature generation.} Features are text attributes that are useful for capturing patterns in data. The most popular features used in machine learning classification are the presence or the frequency of n-grams extracted during the pre-processing step. In the presence-based representation for each instance a binary vector is created in which ``1'' means the presence of a particular n-gram and ``0'' indicates its absence. In the frequency-based representation the number of occurrences of a particular n-gram is used instead of a binary indication of presence. In cases where text length varies greatly, it might be important to use term frequency (TF) and inverse term frequency (IDF) measures \citep{Rajaraman:2011:MMD:2124405}. However, in short messages like tweets words are unlikely to repeat within one instance, making the binary measure of presence as informative as the counts \citep{ikonomakis2005text}.

Apart from the n-grams, additional features can be created to improve the overall quality of text classification. The most common features that are used for this purpose include:

\begin{itemize}
\item Number of words with positive/negative sentiment;
\item Number of negations;
\item Length of a message;
\item Number of exclamation marks;
\item Number of different parts-of-speech in a text (for example, number of nouns, adjectives, verbs);
\item Number of comparative and superlative adjectives.
\end{itemize}

\item \textbf{Feature selection.} Since the main features of a text classifier are N-grams, the dimensionality of the feature space grows proportionally to the size of the dataset. This dramatical growth of the feature space makes it in most cases computationally infeasible to calculate all the features of a sample. Many features are redundant or irrelevant and do not significantly improve the results. Feature selection is the process of identifying a subset of features that have the highest predictive power. This step is crucial for the classification process, since elimination of irrelevant and redundant features allows to reduce the size of feature space increasing the speed of the algorithm, avoiding overfitting as well as contributing to the improved quality of classification.

There are three basic steps in feature selection process \citep{dash1997feature}
\begin{enumerate}
\item \textit{Search procedure}. A process that generates a subset of features for evaluation. A procedure can start with no variables and add them one by one (forward selection) or with all variables and remove one at each step (backward selection), or features can be selected randomly (random selection).
\item \textit{Evaluation procedure}. A process of calculating a score for a selected subset of features. The most common  metrics for evaluation procedure are: Chi-squared, Information Gain, Odds Ratio, Probability Ratio, Document Frequency, Term Frequency. An extensive overview of search and evaluation methods is presented in \citep{ladha2011, Forman:2003:EES:944919.944974}.
\item \textit{Stopping criterion}. The process of feature selection can be stopped based on a: i) search procedure, if a predefined number of features was selected or predefined number of iterations was performed; ii) evaluation procedure, if the change of feature space does not produce a better subset or if optimal subset was found according to the value of evaluation function.
\end{enumerate}

\item \textbf{Learning an Algorithm.} After feature generation and feature selection steps the text is represented in a form that can be used to train an algorithm. Even though many classifiers have been tested for sentiment analysis purposes, the choice of the best algorithm is still not easy since all methods have their advantages and disadvantages (see \citep{marsland2011machine} for more information on classifiers).

\textit{Decision Trees} \citep{Mitchell:1996}. A decision tree text classifier is a tree in which non-leaf nodes represent a conditional test on a feature, branches denote the outcomes of the test, and leafs represent class labels. Decision trees can be easily adapted to classifying textual data and have a number of useful qualities: they are relatively transparent, which makes them simple to understand;  they give direct information about which features are important in making decisions, which is especially true near the top of the decision tree. However, decision trees also have a few disadvantages. One problem is that trees can be easily overfitted. The reason lies in the fact that each branch in the decision tree splits the training data, thus, the amount of training data available to train nodes located in the bottom of the tree, decreases. This problem can be addressed by using the tree pruning. The second weakness of the method is the fact that decision trees require features to be checked in a specific order. This limits the ability of an algorithm to exploit features that are relatively independent of one another.

\textit{Naive Bayes} \citep{NaiveBayesSent} is frequently used for sentiment analysis purposes because of its simplicity and effectiveness. The basic concept of the Naive Bayes classifier is to determine a class (positive negative, neutral) to which a text belongs using probability theory. In case of the sentiment analysis there will be three hypotheses: one for each sentiment class. The hypothesis that has the highest probability will be selected as a class of the text. The potential problem with this approach emerges if some word in the training set appears only in one class and does not appear in any other classes. In this case, the classifier will always classify text to that particular class. To avoid this undesirable effect Laplace smoothing technique may be applied.

Another very popular algorithm is  \textit{Support Vector Machines (SVMs)} \citep{Cortes:1995:SN:218919.218929, Vapnik:1995:NSL:211359}.  For the linearly separable two-class data, the basic idea is to find a hyperplane, that not only separates the documents into classes, but for which the Euclidian distance to the closest training example, or margin, is as large as possible. In a three-class sentiment classification scenario, there will be three pair-wise classifications: positive-negative, negative-neutral, positive-neutral. The method has proved to be very successful for the task of text categorization \citep{Joachims:1999, 288651} since it can handle very well large feature spaces, however, it has low interpretability and is very computationally expensive, because it involves calculations of discretisation, normalization and dot product operations.

\item \textbf{Model Evaluation.} After the model is trained using a classifier it should be validated, typically, using a cross-validation technique, and tested on a hold-out dataset. There are several metrics defined in information retrieval for measuring the effectiveness of classification, among them are:

\begin{itemize}
\item \textit{Accuracy}: as described by \citep{Kotsiantis:2007}, accuracy is ``the fraction of the number of correct predictions over the total number of predictions''.
\item \textit{Error rate}: measures the number of incorrectly predicted instance against the total number of predictions. 
\item \textit{Precision}: shows the proportion of how many instances the model classified correctly to the total number of true positive and true negative examples. In other words, precision shows the exactness of the classifier with respect to each class.
\item \textit{Recall}: represents the proportion of how many instances the model classified correctly to the total number of true positives and false negatives. Recall shows the completeness of the classifier with respect to each class.
\item \textit{F-score}: \citep{Rijsbergen:1979} defined the F1-score as the harmonic mean of precision and recall:
\end{itemize}
  
\begin{equation}
\mbox{\textit{F-Score}} = \frac{2 * Precision * Recall}{Precision + Recall}.
\label{eq:F-Score} 
\end{equation}

Depending on the nature of the task, one may use accuracy, error rate, precision, recall or F-score as a metric or some mixture of them. For example, for unbalanced datasets, it was shown that precision and recall can be better metrics for measuring classifiers performance \citep{manning1999foundations}. However, sometimes one of these metrics can increase at the expense of the other. For example, in the extreme cases the recall can reach to 100\%, but precision can be very low. In these situations the F-score can be a more appropriate measure.
 
\end{enumerate}

    \section{Application of Lexicon and Machine Learning Methods for Twitter Sentiment Classification} \label{sec:Application}
\begin{figure}[t!]
  \centering
	\subfloat[]{\includegraphics[height = 1.7in, width=4.5in]{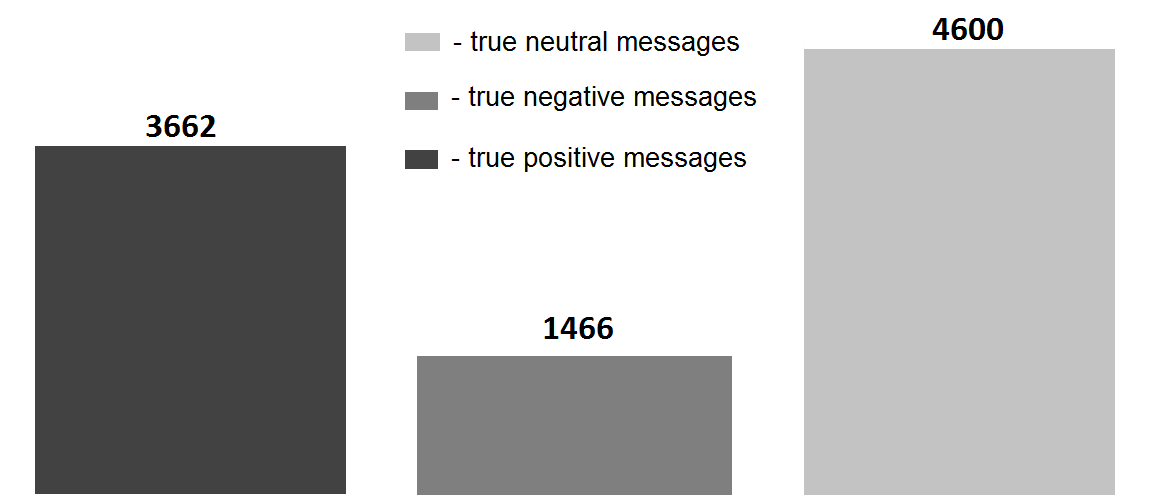}
\label{fig:TrainingData}}
\hfil
\subfloat[]{\includegraphics[height = 1.7in, width=4.5in]{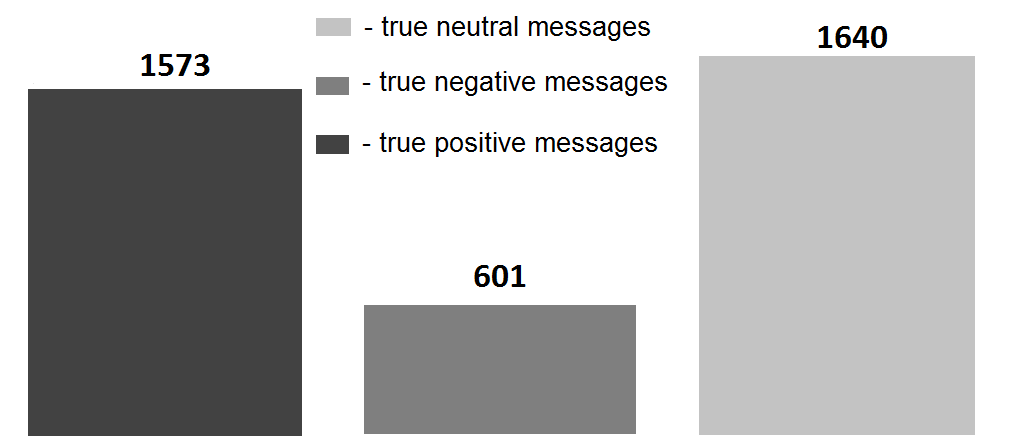}
\label{fig:frog1}}
\caption{Statistics of a) training dataset and b) test datset from SemEval-2013 competition, Task 2-B \citep{SemEval-2013}. Dark grey bar on the left represents the proportion of positive tweets in the dataset,  grey bar in the middle shows the proportion of negative tweets and light grey bar on the right reflects the proportion of neutral sentences.}
\label{fig:SemEvalData}
\end{figure}

Here we provide an example of implementation of the lexicon based approach and the machine learning approach on a case-study. We use benchmark datasets from SemEval-2013 Competition, Task 2: Sentiment Analysis in Twitter, that included two subtasks: A) an expression-level classification, B) a message-level classification \citep{SemEval-2013}. Our interest is in subtask B: ``Given a message, decide whether it is of positive, negative, or neutral sentiment. For messages conveying both a positive and a negative  sentiment, whichever is the stronger one was to be chosen'' \citep{SemEval-2013}. After training and evaluating our algorithm on the training and test datasets provided by SemEval-2013, Task-2 (please, refer to Figure \ref{fig:SemEvalData} for statistics of positive, negative and neutral messages for training and test datasets), we compare our results against the results of 44 participated teams and 149 submissions. 
 
 \FloatBarrier

The second example of application of our algorithm to a large dataset of  42,803,225 Twitter messages related to retail companies is presented in \citep{Souza:2015} and investigates the relationship between Twitter sentiment and stock returns and volatility.

\subsection{Pre-processing}

We performed pre-processing steps as described in section \ref{sec:Preprocessing}. For the most of the  steps we used the machine learning software WEKA\footnote{http://www.cs.waikato.ac.nz/ml/weka/}. WEKA was developed in the university Wakaito and provides implementations of many machine learning algorithms. Since it is an open source tool and has an API, WEKA algorithms can be easily embedded within other applications.

\textbf{Stemming and lemmatisation}. The overall impact of stemming depends on the dataset and stemming algorithm. WEKA contains implementation of a SnowballStemmer \citep{Snowball} and LovinsStemmer \citep{Lovins}. After testing both implementations we discovered that the accuracy of the sentiment classification was decreased after applying both stemming algorithms, therefore, stemming operation was avoided in the final implementation of the sentiment analysis algorithm.

\textbf{Stop-words Removal}. WEKA provides a file with a list of words, which should be considered as stop-words. The file can be adjusted to ones needs. In our study we used a default WEKA stop-list file.

\begin{table}[h]%title of
\caption{Example POS tags.}
\label{tb:POSTable} 
\centering
\begin{tabular}{|p{0.9in}|p{3.3in}|} 
\hline 
@ Tag & Description \\ \hline
@ at-mentions & Is used to identify the user- recipient of the tweet \\ \hline 
U & URL or email address \\ \hline 
\# & Hashtag to identify the topic of the discussion or a category \\ \hline 
\~{} & Discourse marker. Indicates, that message is the continuation of the previous tweet  \\ \hline 
E & Emoticons , , etc. \\ \hline 
G & Abbreviations, shortenings of  words \\ \hline 
\end{tabular}
\end{table}

\begin{table}[h]
\caption{Example of ArkTweetNLP  \citep{Gimpel:2011:PTT:2002736.2002747} tagger in practice.} 
\label{tb:ArkTweetNLP}
\centering
\begin{tabular}{|p{4.5in}|} \hline 
\textbf{Sentence:\newline ikr smh he asked fir yo last name so he can add u on fb lololol\newline }\underbar{word}    \underbar{tag}     \underbar{confidence}\newline ikr        !       0.8143\newline smh     G       0.9406\newline he        O       0.9963\newline asked   V       0.9979\newline fir        P       0.5545\newline yo       D       0.6272\newline last      A       0.9871\newline name   N       0.9998\newline so        P       0.9838\newline he       O       0.9981\newline can      V       0.9997\newline add     V       0.9997\newline u         O       0.9978\newline on       P       0.9426\newline fb        \^{}       0.9453\newline lololol  !       0.9664\newline \newline``ikr'' means ``I know, right?'', tagged as an interjection. \newline ``so'' is being used as a subordinating conjunction, which our coarse tagset denotes P. \newline ``fb'' means ``Facebook'', a very common proper noun (\^{}). \newline ``yo'' is being used as equivalent to ``your''; our coarse tagset has posessive pronouns as D. ``fir'' is a misspelling or spelling variant of the preposition for.\textbf{\newline  }Perhaps the only debatable errors in this example are for ikr and smh (``shake my head''): should they be G for miscellaneous acronym, or ! for interjection?\textbf{} \\ \hline 
\end{tabular}
\end{table}

\textbf{Part-of-Speech Tagging (POS).} In the current study we tested performance of multiple existing pos-taggers: Stanford Tagger\footnote{http://nlp.stanford.edu/software/index.shtml}, Illinois Tagger\footnote{http://cogcomp.cs.illinois.edu/page/software\_view/3 }, OpenNLP\footnote{ http://opennlp.sourceforge.net/models-1.5}, LingPipe POS Tagger\footnote{http://alias-i.com/lingpipe/demos/tutorial/posTags/read-me.html}, Unsupos\footnote{http://wortschatz.uni-leipzig.de/\~{}cbiemann/software/unsupos.html}, ArkTweetNLP\footnote{http://www.ark.cs.cmu.edu/TweetNLP}, Berkeley NLP Group Tagger\footnote{http://nlp.cs.berkeley.edu/Software.shtml}. We finally chose to use ArkTweetNLP library developed by the team of researchers from Carnegie Mellon University \citep{Gimpel:2011:PTT:2002736.2002747} since it was trained on a Twitter dataset. ArkTweetNLP developed 25 POS tags, with some of them specifically designed for special Twitter symbols, such as hashtags, at-mentions, retweets, emoticons, commonly used abbreviations (see Table \ref{tb:POSTable} for some tags examples). An example\footnote{http://www.ark.cs.cmu.edu/TweetNLP/} of how ArkTweetNLP tagger works in practice is presented in Table \ref{tb:ArkTweetNLP}.

As the result of POS-tagging in our study, we filtered out all words that did not belong to one of the following categories: N(common noun), V(verb), A(adjective), R(adverb), !(interjection), E(emoticon), G(abbreviations, foreign words, possessive endings).

\textbf{Negations Handling.} We implemented negation handling using simple, but effective strategy: if negation word was found, the sentiment score of every word appearing between a negation and a clause-level punctuation mark (.,!?:;) was reversed \citep{Pang:2002:TUS:1118693.1118704}. There are, however, some grammatical constructions in which a negation term does not have a scope. Some of these situations we implemented as exceptions:

\textit{Exception Situation 1:} Whenever a negation term is a part of a phrase that does not carry negation sense, we consider that the scope for negation is absent and the polarity of words is not reversed. Examples of these special phrases include ``not only'', ``not just'', ``no question'', ``not to mention'' and ``no wonder''.

\textit{Exception Situation 2:} A negation term does not have a scope when it occurs in a negative rhetorical question. A negative rhetorical question is identified by the following heuristic. (1) It is a question; and (2) it has a negation term within the first three words of the question. For example:\newline
    ``Did not I enjoy it?''\newline
    ``Wouldn't you like going to the cinema?''
   
\textbf{Tokenisation into N-grams.} We used WEKA tokeniser to extract uni-grams and bi-grams from the Twitter dataset.

\subsection{Lexicon Approach}

\subsubsection{Automatic Lexicon Generation.}
In this study we aimed to create a lexicon specifically oriented for sentiment analysis of Twitter messages. For this purpose we used the approach described in \ref{sec:Lexicon}: ``Constructing a lexicon from trained data'' and the training dataset from Mark Hall \citep{MHall} that is comprised of manually labelled 41403 positive Twitter messages and 8552 negative Twitter messages. The method to generate a sentiment lexicon was implemented as follows: 

\begin{enumerate}
\item  Pre-processing of the dataset: POS tags were assigned to all words in the dataset; words were lowered in case; BOW was created by tokenising the sentences in the dataset.

\item  The number of occurrences of each word in positive and negative sentences from the training dataset was calculated.

\item  The positive polarity of each word was calculated by dividing the number of occurrences in positive sentences by the number of all occurrences:

\begin{equation}
positiveSentScore=\ \frac{\#Positive\ sentences}{\left(\#Positive\ sentences+\#Negative\ sentences\right)}.
\end{equation}

For example, we calculated that the word ``\textit{pleasant}'' appeared 122 times in the positive sentences and 44 times in the negative sentences. According to the formula, the positive sentiment score of the word ``\textit{pleasant}'' is

\[positiveSentScore=\ \frac{122\ }{\left(122+44\right)}=\ 0.73.\]

Similarly, the negative score for the word ``\textit{pleasant}'' can be calculated by dividing the number of occurrences in negative sentences by the total number of mentions

\begin{equation}
negativeSentScore=\ \frac{\#Negative\ sentences}{\left(\#Positive\ sentences+\#Negative\ sentences\right)},
\label{eq:SentScore2} 
\end{equation}

\[negativeSentScore=\ \frac{44\ }{\left(122+44\right)}=\ 0.27.\]

Based on the positive score of the word we can make a decision about its polarity: the word is considered positive, if its positive score is above 0.6; the word is considered neutral, if its positive score is in the range [0.4; 0.6]; the word is considered negative, if the positive score is below 0.4. Since the positive score of the word ``pleasant'' is 0.73, it is considered to carry positive sentiment. Sentiment scores of some other words from the experiment are presented in Table \ref{tb:sentScores}.

\begin{table}[h]
\caption{Example of sentiment scores of words in the automatically generated lexicon.}
\centering
\begin{tabular}{|p{0.6in}|p{0.5in}|p{0.5in}|p{0.5in}|} \hline 
 & GOOD & BAD & LIKE \\ \hline 
Positive Score & 0.675 & 0.213 & 0.457  \\ \hline 
Negative Score & 0.325 & 0.787 &0.543 \\ \hline 
\end{tabular}
\label{tb:sentScores}
\end{table}

We can observe from the table that the words ``\textit{GOOD}'' and ``\textit{BAD}'' have strongly defined positive and negative scores, as we would expect. The word ``\textit{LIKE}'' has polarity scores ranging between 0.4 and 0.6 indicating its neutrality. To understand why the ``neutral'' label for the word ``\textit{LIKE}'' was assigned we investigate the semantic role of this word in English language: 
 
\begin{enumerate}
\item  Being a verb to express preference. For example: ``\textit{I like ice-cream}''.
\item  Being a preposition for the purpose of comparison. For example: ``\textit{This town looks like Brighton.}''
\end{enumerate}

The first sentence has positive sentiment, however can easily be transformed into a negative sentence: ``\textit{I don't like ice-cream}''. This demonstrates that the word ``\textit{LIKE}'' can be used with equal frequency for expressing positive and negative opinions. In the second example the word ``\textit{LIKE}'' is playing a role of a preposition and does not effect the overall polarity of the sentence. Thus, the word ``\textit{LIKE}'' is a neutral word and was correctly assigned a neutral label using the approach described above.

In our study all words from the Bag-of-Words with a polarity in the range [0.4; 0.6] were removed, since they do not help to classify the text as positive or negative. The sentiment scores of the words were mapped into the range [-1;1] by using the following formula:

\begin{equation}
PolarityScore = 2*positiveSentScore - 1 .
\label{eq:H2}
\end{equation}

According to this formula, the word ``LIKE'', obtained a score 0.446 * 2 -1 = -0.1, which indicates the neutrality of the word. In case when the word is extremely positive and had a $positiveSentScore$ of 1, the mapped score will be positive:  1 * 2 -- 1 = 1.  If the word is extremely negative and has the $positiveSentScore$ equal to 0, the mapped score will be negative: 0 * 2 - 1 = -1.

\end{enumerate}

\subsubsection{Lexicons Combinations.}

Since the role of emoticons for expressing opinion online is continuously increasing, it is crucial to incorporate emoticons into lexicons used for sentiment analysis. Hogenboom et al. showed that incorporation of the emoticons into lexicon can significantly improve the accuracy of classification \citep{Hogenboom:2013:EES:2480362.2480498}. Apart from emoticons, new slang words and abbreviations are constantly emerging and need to be accounted for when performing sentiment analysis. However, most of the existing public lexicons do not contain emoticons and social-media slang, on the contrary, emoticons and abbreviations are often being removed as typographical symbols during the first stages of pre-processing. 

\begin{table}[!h]
\caption{Example of tokens from our EMO lexicon along with their polarity. Tokens represent emoticons, abbreviations and slang words that are used in social-media to express emotions.}
\centering
\begin{tabular}{|p{0.5in}|p{0.3in}|p{0.5in}|p{0.3in}|p{0.7in}|p{0.3in}|p{0.7in}|p{0.3in}|} \hline 
\textbf{Emoticon} & \textbf{Score} & \textbf{Emoticon} & \textbf{Score} & \textbf{Abbreviation} & \textbf{Score} & \textbf{Abbreviation} & \textbf{Score}\\ \hline 
l-) & 1 & [-( & -1 & lol & 1 & dbeyr & -1 \\ \hline 
:-\} & 1 & T\_T & -1 & ilum & 1 & iwiam & -1\\ \hline 
x-d & 1 & :-(( & -1 & iyqkewl & 1 & nfs & -1\\ \hline
;;-) & 1 & :-[ & -1 & iwalu & 1 & h8ttu & -1\\ \hline 
=] & 1 & :((( & -1 & koc & 1 & gtfo & -1\\ \hline
\end{tabular}
\label{fig:EMOLexicon}
\end{table}  \FloatBarrier

In this study we manually constructed a lexicon of emoticons, abbreviations and slang words commonly used in social-media to express emotions (EMO). Example of tokens from our lexicon are presented in Table \ref{fig:EMOLexicon}. We aimed to analyse how performance of the classic opinion lexicon (OL) \citep{BingLexicon}  can be improved by enhancing it with our EMO lexicon. We also expanded the lexicon further by incorporating words from the automatically created lexicon (AUTO). The process of automatic lexicon creation was described in detail in the previous section.

With opinion lexicon (OL) serving as a baseline, we compared the performance of some lexicon combinations as shown in Table \ref{tb:Combinations}:

\begin{table}[!h]
\caption{Combinations of lexicons tested} %title of the table
\label{tb:Combinations}
\centering
\begin{tabular}{|p{0.3in}|p{1.7in}|} 
\hline
%inserting double
 & Lexicons combinations\\
\hline 
1. & OL\\ 
2. & OL + EMO\\
3. & OL + EMO + AUTO\\
\hline
\end{tabular}
\end{table} 

\subsubsection{Sentiment Score Calculation.}
    
In this study we calculate sentiment scores of tweets as described in section \ref{sec:sent} using Equation \ref{eq:LexiconScore}. We also propose an alternative measure based on the logarithm of the standard score. We  normalise the logarithmic score in such a way that the values range between [-1; 1] with -1 being the most negative score and 1 being the most positive score (see \ref{eq:LogScore}).

\begin{equation}
Score_{Log{10}} = 
 \begin{cases}
{sign}({Score}_{AVG})Log_{10}(\mid 10{Score}_{AVG}\mid), & \text{if } \mid{Score}_{AVG}\mid > 0.1 , \\
0, & \text{otherwise} \label{eq:LogScore}
\end{cases}
\end{equation}

\subsubsection{Lexicon Performance Results.}

\begin{figure}[h] 
%\begin{wrapfigure}{h}{0.3\textwidth}
\centering
\includegraphics[width=1.0\textwidth]{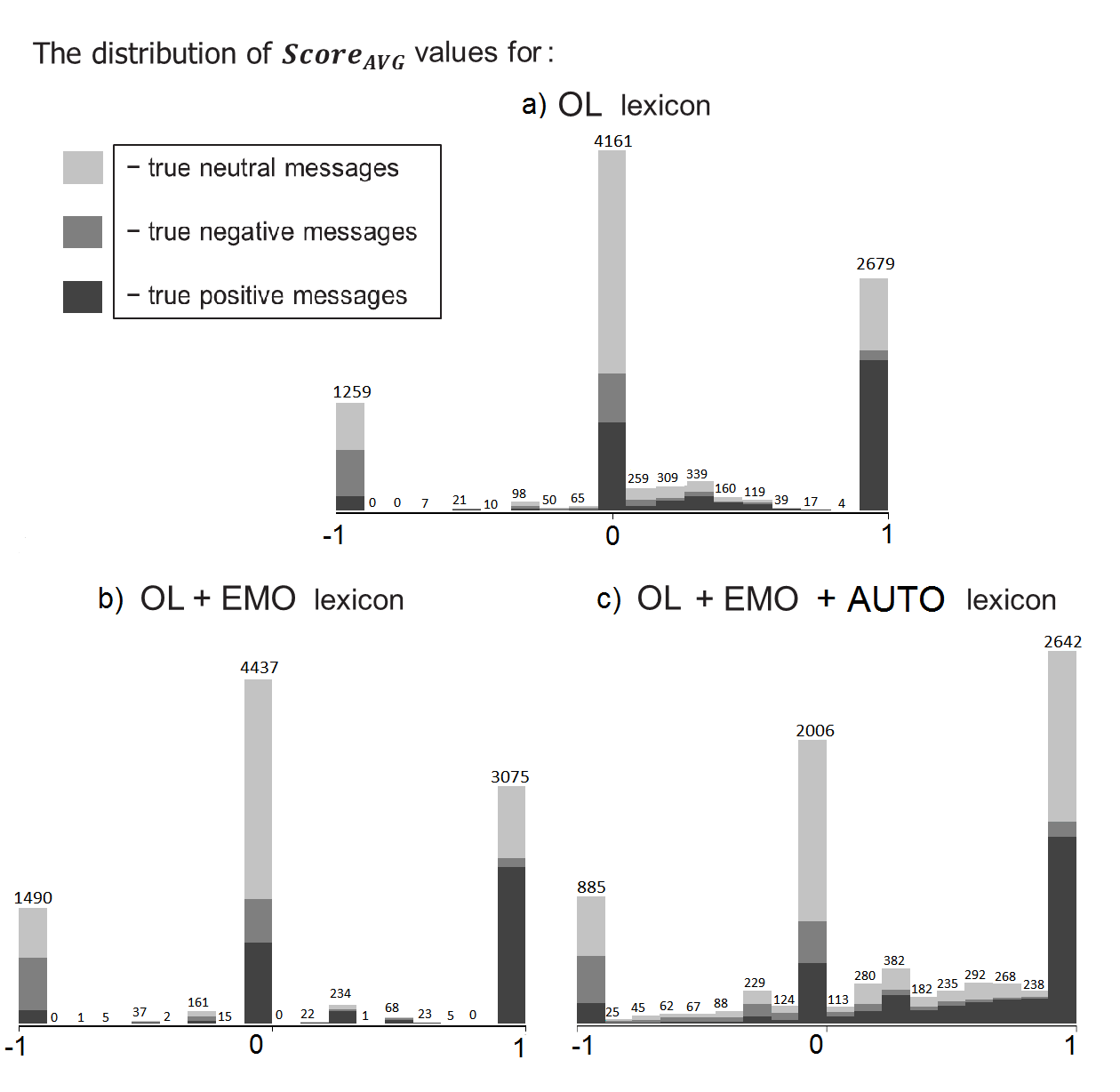}
\caption{Histograms of sentiment scores for different lexicon combinations using the Simple Average Score. The colors of the bars represent the true labels of the tweets: dark grey stands for positive messages, light grey for neutral messages and medium grey stands for positive messages.}
\label{fig:AVGScores}
\end{figure}

The analysis of performance of our algorithm was conducted on the test dataset from SemEval-2013, Task 2-B \citep{SemEval-2013} (see Figure \ref{fig:frog1}). Figure \ref{fig:AVGScores} presents the results for the three different lexicons using the \textit{Simple Average} as the sentiment score (Equation \ref{eq:LexiconScore}). The values of the sentiment score range from -1 to 1. The colors of the bars represent the true labels of the tweets: dark grey stands for positive messages, light grey for neutral messages and medium grey stands for negative messages. In the case of perfect classification, we would obtain a clear separation of the colors. However, from Figure \ref{fig:AVGScores} we can see that classification for all three lexicons was not ideal. For example, all lexicons made the biggest mistake in misclassifying neutral messages (we can see that light grey color is present for the sentiment scores of -1 and 1 in all three histograms, indicating that some of neutral messages were classified as positive or negative). This phenomenon can be explained with the fact that even neutral messages often contain one or more polarity words, which leads to the final score of the message being a value different from 0 and being classified as positive or negative.

\begin{figure}[h] 
%\begin{wrapfigure}{h}{0.3\textwidth}
\centering
\includegraphics[width=1.0\textwidth]{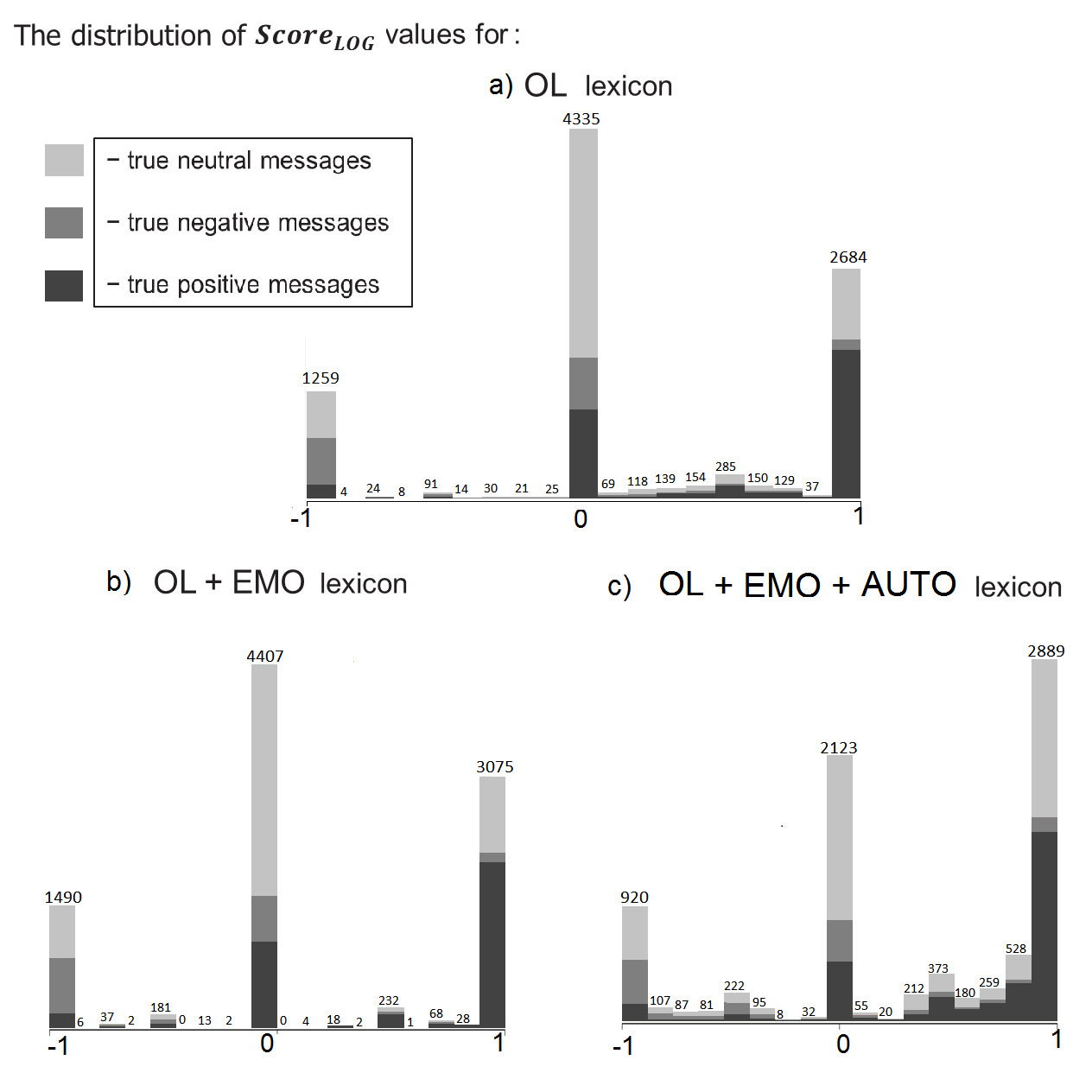}
\caption{Histograms of sentiment scores for different lexicon combinations using the Logarithmic Score. The colors of the bars represent the true labels of the tweets: dark grey stands for positive messages, light grey for neutral messages and medium grey stands for positive messages.}
\label{fig:LOGScores}
\end{figure}

The results based on the logarithmic approach (Equation \ref{eq:LogScore}) reveal that positive, negative and neutral classes became more defined (Figure \ref{fig:LOGScores}). Indeed, the logarithmic score makes it easier to set up the thresholds for assigning labels to different classes, thus, we can conclude that using a logarithmic score for calculating sentiment is more appropriate than using a simple average score. 

To compare the performance of three lexicon combinations we need to assign positive, negative or neutral labels to the tweets based on the calculated sentiment scores, and compare the predicted labels against the true labels of tweets. For this purpose we employ a k-means clustering algorithm, using Simple Average and Logarithmic scores as features. The results of K-means clustering for the 3 lexicons and 2 types of sentiment scores are reported in Table \ref{tb:LexiconCombinations}.

\begin{table}[!h]
\caption{Results of K-Means clustering for different lexicon combinations.}
\centering
\begin{tabular}{|p{0.8in}|p{0.8in}|p{0.8in}|p{1.2in}|} \hline 
\textbf{Accuracy} & \textbf{OL} & \textbf{OL + EMO} & \textbf{OL + EMO + AUTO} \\ \hline 
\textbf{${Score}_{AVG}$} & 57.07\% & 60.12\% & 51.33\%  \\ \hline 
\textbf{$Score_{Log{10}}$} & 58.43\% & 61.74\% & 52.38\%  \\ \hline 
\end{tabular}
\label{tb:LexiconCombinations} 
\end{table}  \FloatBarrier

As shown in Table \ref{tb:LexiconCombinations} the lowest accuracy of classification for both types of scores corresponded to the biggest lexicon (OL + EMO + AUTO). This result can be related to a noisy nature of Twitter data. Training a lexicon on noisy data could have introduced ambiguity regarding the sentiment of individual words. Thus, automatic generation of the lexicon (AUTO) based on Twitter labelled data cannot be considered a reliable technique. The small OL lexicon showed better results since it consisted mainly of adjectives that carry strong positive or negative sentiment that are unlikely to cause ambiguity. The highest accuracy of classification 61.74\% was achieved using the combination of OL and EMO lexicons (OL + EMO) and a logarithmic score. This result confirms that enhancing the lexicon for Twitter sentiment analysis with emoticons, abbreviations and slang words increases the accuracy of classification. It is important to notice that the Logarithmic Score provided an improvement of 1.36\% over the Simple Average Score.

\subsection{Machine Learning Approach}

We performed Machine Learning based sentiment analysis. For this purpose we used the machine learning package WEKA\footnote{http://www.cs.waikato.ac.nz/ml/weka/}.

 \subsubsection{Pre-processing/cleaning the data.}
    
Before training the classifiers the data needed to be pre-processed and this step was performed according to the general process described in section \ref{sec:Preprocessing}. Some additional steps that had to be performed:

\begin{itemize}

\item \textit{Filtering.} Some syntactic constructions used in Twitter messages are not useful for sentiment detection. These constructions include URLs, @-mentions, hashtags, RT-symbols and they were removed during the pre-processing step.
\item \textit{Tokens replacements.} The words that appeared to be under the effect of the negation words were modified by adding a suffix \textbf{\_NEG} to the end of those words.

For example, the phrase \textit{I don't want.} was modified to \textit{I don't want\_NEG}.

This modification is important, since each word in a sentence serves a purpose of a feature during the classification step. Words with  \textbf{\_NEG} suffixes increase the dimensionality of the feature space, but allow the classifier to distinguish between words used in the positive and in the negative context.

When performing tokenisation, the symbols ():;, among others are considered to be delimiters, thus most of the emoticons could be lost after tokenisation. To avoid this problem positive emoticons were replaced with \textbf{pos\_emo} and negative were replaced with \textbf{neg\_emo}. Since there are many variations of emoticons representing the same emotions depending on the language and community, the replacement of all positive lexicons by ‘pos\_emo’ and all negative emoticons by ‘neg\_emo’ also achieved the goal of significantly reducing the number of features.

\end{itemize}

 \subsubsection{Feature Generation.} 

The following features were constructed for the purpose of training a classifier:

\begin{itemize}

\item \textbf{N-grams:} we transformed the training dataset into the bag-of-ngrams taking into account only the presence/absence of unigrams. Using ``n-grams frequency'' would not be logical in this particular experiment, since Twitter messages are very short, and a term is unlikely to appear in the same message more than once;
\item \textbf{Lexicon Sentiment:} the sentiment score obtained during the lexicon based sentiment analysis as described in \ref{eq:LogScore};
\item \textbf{Elongated words number}: the number of words with one character repeated more than 2 times, e.g. 'soooo';
\item \textbf{Emoticons}: presence/absence of positive and negative emoticons at any position in the tweet; 
\item \textbf{Last token}: whether the last token is a positive or negative emoticon;
\item \textbf{Negation}: the number of negated contexts;
\item \textbf{POS}: the number of occurrences for each part-of-speech tag: verbs, nouns, adverbs, at-mentions, abbreviations, URLs, adjectives and others
\item \textbf{Punctuation marks}: the number of occurrences of punctuation marks in a tweet;
\item \textbf{Emoticons number}: the number of occurrences of positive and negative emoticons;
\item \textbf{Negative tokens number}: total count of tokens in the tweet with logarithmic score less than 0;
\item \textbf{Positive tokens number}: total count of tokens in the tweet with logarithmic score greater than 0;

\end{itemize}

 \subsubsection{Feature Selection.}
After performing the feature generation step described above a feature space comprising \textbf{1826} features was produced. The next important step for improving classification accuracy is the selection of the most relevant features from this feature space. To this purpose we used Information Gain evaluation algorithm  and a Ranker search method \citep{Ladha-featuresSelection}. Information Gain measures the decrease in entropy when the feature is present vs absent, while Ranker ranks the features based on the amount of reduction in the objective function. We used features for which the value of information gain was above zero. As the result, a subset of \textbf{528} features was selected. 

\begin{table}[!h]
\caption{Example of top selected features.}
\centering
\begin{tabular}{|p{1.0in}|p{1.0in}|p{1.0in}|p{1.0in}|} \hline 
\textbf{TOP FEATURES} & 11. great & 22. fun & 33. hope \\ \hline 
1. LexiconScore & 12. posV & 23. lastTokenScore & 34. thanks\\ \hline 
2. maxScore & 13. happy & 24. i love & 35. luck\\ \hline 
3. posR & 14. love & 25. don & 36. best\\ \hline
4. minScore & 15. excited & 26. don't & 37. i don't\\ \hline 
5. negTokens & 16. can't & 27. amazing & 38. looking forward\\ \hline
6. good & 17. i & 28. fuck & 39. sorry\\ \hline 
7. posE & 18. not & 29 love you & 40. didn't\\ \hline 
8. posN & 19. posA & 30. can & 41. hate\\ \hline 
9. posU & 20. posElongWords & 31. awesome & 42. ... \\ \hline 
\end{tabular}
\label{tb:featuresSelection}
\end{table}  \FloatBarrier

Some of the  top selected features are displayed in Table \ref{tb:featuresSelection}, revealing that the ``Lexicon Sentiment'' feature, described in the previous section as a ``Lexicon Sentiment'', is located at the top of the list. This important result demonstrates that the ``Lexicon Sentiment'' plays a leading role in determining the final sentiment polarity of the sentence. Other highly ranked features included: minimal and maximum scores, number of negated tokens, number of different parts-of-speech in the message. To validate the importance of the ``Lexicon Sentiment'' feature and other manually constructed features, we performed cross-validation tests according to two scenarios: i) in the first scenario (Table \ref{tb:Movies1}) we trained three different classifiers using only N-grams as features; ii) in the second scenario (Table \ref{tb:Movies2}) we trained the models using traditional N-grams features in combination with the ``Lexicon Sentiment'' feature and other manually constructed features: number of different parts-of-speech, number of emoticons, number of elongated words. Tests were performed on a movie review dataset ``Sentence Polarity Dataset v 1.0 ''\footnote{http://www.cs.cornell.edu/people/pabo/movie-review-data/} released by Bo Pang and Lillian Lee in 2005 and comprised of 5331 positive and 5331 negative processed sentences.

As it can be observed from tables \ref{tb:Movies1} and \ref{tb:Movies2}, the addition of the ``Lexicon Sentiment'' feature and other manually constructed features allowed to increase all performance measures significantly for 3 classifies. For example, the accuracy of Naive Bayes classifier was increased by 7\%,  accuracy of Decision Trees was increased by over 9\%, and the accuracy of SVM improved by 4.5\%.

\begin{table}[!h]
\caption{Scenario 1: 5-fold cross-validation test on a movies reviews dataset using only N-grams as features.}
\centering
\begin{tabular}{|p{0.6in}|p{0.6in}|p{0.6in}|p{0.5in}|p{0.5in}|p{0.5in}|p{0.5in}|} \hline  
\textbf{Method} & \textbf{Tokens Type} & \textbf{Folds Number} & \textbf{Accuracy} & \textbf{Precision} & \textbf{Recall} & \textbf{F-Score}\\ \hline 
\textbf{Naive Bayes}  & uni/bigrams  & 5 & 81.5\% & 0.82 & 0.82 & 0.82 \\ \hline 
\textbf{Decision Trees} & uni/bigrams  & 5 & 80.57\% & 0.81 & 0.81 & 0.81 \\ \hline 
\textbf{SVM} & uni/bigrams  & 5 & 86.62\% & 0.87 & 0.87 & 0.87 \\ \hline 
\end{tabular}
\label{tb:Movies1} 
\end{table}  \FloatBarrier

\begin{table}[!h]
\caption{Scenario 2: 5-fold cross-validation test on a movies reviews dataset using traditional N-grams features in combination with manually constructed features: lexicon sentiment score, number of different parts-of-speech, number of emoticons, number of elongated words, etc.}
\centering
\begin{tabular}{|p{0.6in}|p{0.6in}|p{0.6in}|p{0.5in}|p{0.5in}|p{0.5in}|p{0.5in}|} \hline  
\textbf{Method} & \textbf{Tokens Type} & \textbf{Folds Number} & \textbf{Accuracy} & \textbf{Precision} & \textbf{Recall} & \textbf{F-Score}\\ \hline 
\textbf{Naive Bayes}  & uni/bigrams  & 5 & 88.54\% & 0.89 & 0.86 & 0.86 \\ \hline 
\textbf{Decision Trees} & uni/bigrams  & 5 & 89.9\% & 0.90 & 0.90 & 0.90 \\ \hline 
\textbf{SVM} & uni/bigrams  & 5 & 91.17\% & 0.91 & 0.91 & 0.91 \\ \hline 
\end{tabular}
\label{tb:Movies2} 
\end{table}  \FloatBarrier

 \subsubsection{Training the Model, Validation and Testing.}

Machine Learning Supervised approach requires a labelled training dataset. We used a publicly available training dataset (Figure \ref{fig:TrainingData}) from SemEval-2013 competition, Task 2-B \citep{SemEval-2013}. 

Each of the tweets from the training set was expressed in terms of its attributes. As the result, $n$ by $m$ binary matrix was created, where $n$ is the number of training instances and $m$ is the number of features. This matrix was used for training different classifiers: Naive Bayes, Support Vector Machines, Decision trees. It is important to notice that the training dataset was highly unbalanced with the majority of neutral messages (Figure \ref{fig:TrainingData}). In order to account for this unbalance we trained a cost-sensitive SVM model \citep{LingS2007}. Cost-Sensitive classifier allows to minimize the total cost of classification by putting a higher cost on a particular type of error (in our case, misclassifying positive and negative messages as neutral). 

As the next step we tested the models on an unseen before test set (Figure \ref{fig:frog1}) from SemEval-2013 Competition \citep{SemEval-2013} and compared our results against the results of 44 teams that took part in the SemEval-2013 competition. While the classification was performed for 3 classes (pos, neg, neutral), the evaluation metric was F-score (Equation \ref{eq:F-Score}) between positive and negative classes.

\begin{table}[!h]
\caption{F-score results of our algorithm using different classifiers. The test was performed on a test dataset from SemEval Competition-2013, Task 2-B \citep{SemEval-2013}.}
\centering
\begin{tabular}{|p{0.7in}|p{0.6in}|p{0.6in}|p{0.6in}|p{1.0in}|} \hline 
\textbf{Classifier} & \textbf{Naive Bayes} & \textbf{Decision Trees} & \textbf{SVM} & \textbf{Cost Sensitive SVM}\\ \hline 
\textbf{F-SCORE}  & 0.64  & 0.62   & 0.66   & 0.73   \\ \hline 
\end{tabular}
\label{tb:Classifiers} 
\end{table}  \FloatBarrier

\begin{table}[!h]
\caption{Fscore results of SemEval Competition-2013, Task 2-B \citep{SemEval-2013}.}
\centering
\begin{tabular}{|p{1.6in}|p{1.5in}|} \hline 
\textbf{TEAM NAME} & \textbf{F-SCORE} \\ \hline 
NRC-Canada & 0.6902  \\ \hline 
GUMLTLT & 0.6527 \\ \hline 
TEREGRAM & 0.6486 \\ \hline 
AVAYA  &  \\ \hline 
BOUNCE & 0.6353 \\ \hline 
KLUE & 0.6306  \\ \hline 
AMI and ERIC & 0.6255 \\ \hline 
FBM & 0.6117 \\ \hline 
SAIL  &  \\ \hline 
AVAYA & 0.6084  \\ \hline 
SAIL & 0.6014   \\ \hline 
UT-DB & 0.5987  \\ \hline 
FBK-irst & 0.5976    \\ \hline 
\end{tabular}
\label{tb:MLResults} 
\end{table}  \FloatBarrier

Our results for different classifiers are presented in Table \ref{tb:Classifiers}. We can observe that the Decision Tree algorithm had the lowest F-score of 62\%. The reason may lay in a big size of the tree needed to incorporate all of the features. Because of the tree size, the algorithm needs to traverse multiple nodes until it reaches the leaf and predicts the class of the instance. This long path increases the probability of mistakes and thus decreases the accuracy of the classifier. Naive Bayes and SVM produced better scores of 64\% and 66\% respectively. The best model was a Cost-sensitive SVM that allowed to achieve the F-measure of 73\%. This is an important result, providing evidence that accounting for the unbalance in the training dataset allows to improve model performance significantly. Comparing our results with the results of the competition (Table \ref{tb:MLResults}), we can conclude that our algorithm based on the Cost-sensitive SVM would had produced the best results scoring 4 points higher than the winner of that competition.